\newif\ifeusipstyle
\newif\ifdohybrid
\eusipstylefalse
\dohybridfalse

\ifeusipstyle
  \documentclass[a4paper]{article}
  \usepackage{spconf,amsmath,graphicx,cite}
\else
  \documentclass[conference]{IEEEtran} 

  \usepackage[dvips]{graphicx}
\fi

\usepackage{url}
\usepackage{amsfonts}

\title{Maximum Entropy Auto-Encoding}

\ifeusipstyle
   \name{Paul M Baggenstoss}
   \address{Fraunhofer FKIE, Fraunhoferstr 20,
   \\ 53343 Wachtberg, Germany}
\else
   \author{\IEEEauthorblockN{Paul M. Baggenstoss}
   \IEEEauthorblockA{Fraunhofer FKIE,
   Fraunhoferstrasse 20\\
   53343 Wachtberg, Germany\\
   Email: p.m.baggenstoss@ieee.org}
   }
\fi

\begin{document}
\newcommand{\defined}{\stackrel{\mbox{\tiny$\Delta$}}{=}}
\newtheorem{example}{Example}
\newtheorem{conclusion}{Conclusion}
\newtheorem{assumption}{Assumption}
\newtheorem{definition}{Definition}
\newtheorem{problem}{Problem}
\newcommand{\erf}{{\rm erf}}

\newcommand{\sst}{\scriptstyle }
\newcommand{\xparen}{\mbox{\small$(\bfx)$}}
\newcommand{\hojz}{H_{0j}\mbox{\small$(\bfz)$}}
\newcommand{\Hozj}{H_{0,j}\mbox{\small$(\bfz_j)$}}
\newcommand{\smallmath}[1]{{\scriptstyle #1}}
\newcommand{\Hoz}[1]{H_0\mbox{\small$(#1)$}}
\newcommand{\Hozp}[1]{H_0^\prime\mbox{\small$(#1)$}}
\newcommand{\Hozpp}[1]{H_0^{\prime\prime}\mbox{\small$(#1)$}}
\newcommand{\hoz}{\Hoz{\bfz}}
\newcommand{\hooz}{\Hozp{\bfz}}
\newcommand{\hoooz}{\Hozpp{\bfz}}
\newcommand{\smJ}{{\scriptscriptstyle \! J}}
\newcommand{\smK}{{\scriptscriptstyle \! K}}

\newcommand{\erfc}{{\rm erfc}}
\newcommand{\bitem}{\begin{itemize}}
\newcommand{\dsum}{{ \displaystyle \sum}}
\newcommand{\eitem}{\end{itemize}}
\newcommand{\benum}{\begin{enumerate}}
\newcommand{\eenum}{\end{enumerate}}
\newcommand{\bdm}{\begin{displaymath}}
\newcommand{\bfzro}{{\underline{\bf 0}}}
\newcommand{\bfone}{{\underline{\bf 1}}}
\newcommand{\edm}{\end{displaymath}}
\newcommand{\beq}{\begin{equation}}
\newcommand{\bea}{\begin{eqnarray}}
\newcommand{\eea}{\end{eqnarray}}
\newcommand{\cali}{ {\cal \bf I}}
\newcommand{\caln}{ {\cal \bf N}}
\newcommand{\barray}{\begin{displaymath} \begin{array}{rcl}}
\newcommand{\earray}{\end{array}\end{displaymath}}
\newcommand{\eeq}{\end{equation}}
\newcommand{\qed}{\framebox{$\;$}}
\newcommand{\btheta}{\mbox{\boldmath $\theta$}}
\newcommand{\bTheta}{\mbox{\boldmath $\Theta$}}
\newcommand{\blam}{\mbox{\boldmath $\Lambda$}}
\newcommand{\bdelta}{\mbox{\boldmath $\delta$}}
\newcommand{\bgamma}{\mbox{\boldmath $\gamma$}}
\newcommand{\balpha}{\mbox{\boldmath $\alpha$}}
\newcommand{\bbeta}{\mbox{\boldmath $\beta$}}
\newcommand{\balphascript}{\mbox{\boldmath ${\scriptstyle \alpha}$}}
\newcommand{\bbetascript}{\mbox{\boldmath ${\scriptstyle \beta}$}}
\newcommand{\bLambda}{\mbox{\boldmath $\Lambda$}}
\newcommand{\bDelta}{\mbox{\boldmath $\Delta$}}
\newcommand{\bomega}{\mbox{\boldmath $\omega$}}
\newcommand{\bOmega}{\mbox{\boldmath $\Omega$}}
\newcommand{\blambda}{\mbox{\boldmath $\lambda$}}
\newcommand{\bphi}{\mbox{\boldmath $\phi$}}
\newcommand{\bpi}{\mbox{\boldmath $\pi$}}
\newcommand{\bnu}{\mbox{\boldmath $\nu$}}
\newcommand{\brho}{\mbox{\boldmath $\rho$}}
\newcommand{\bmu}{\mbox{\boldmath $\mu$}}
\newcommand{\sigi}{\mbox{\boldmath $\Sigma$}_i}
\newcommand{\bfu}{{\bf u}}
\newcommand{\bfx}{{\bf x}}
\newcommand{\bfb}{{\bf b}}
\newcommand{\bfk}{{\bf k}}
\newcommand{\bfc}{{\bf c}}
\newcommand{\bfv}{{\bf v}}
\newcommand{\bfn}{{\bf n}}
\newcommand{\bfK}{{\bf K}}
\newcommand{\bfh}{{\bf h}}
\newcommand{\bff}{{\bf f}}
\newcommand{\bfg}{{\bf g}}
\newcommand{\bfe}{{\bf e}}
\newcommand{\bfr}{{\bf r}}
\newcommand{\bfw}{{\bf w}}
\newcommand{\calX}{{\cal X}}
\newcommand{\calZ}{{\cal Z}}
\newcommand{\bx}{{\bf x}}
\newcommand{\bb}{{\bf b}}
\newcommand{\by}{{\bf y}}
\newcommand{\bfy}{{\bf y}}
\newcommand{\bfz}{{\bf z}}
\newcommand{\bfs}{{\bf s}}
\newcommand{\bfa}{{\bf a}}
\newcommand{\bfA}{{\bf A}}
\newcommand{\bfB}{{\bf B}}
\newcommand{\bfV}{{\bf V}}
\newcommand{\bfZ}{{\bf Z}}
\newcommand{\bfH}{{\bf H}}
\newcommand{\bfX}{{\bf X}}
\newcommand{\bfR}{{\bf R}}
\newcommand{\bfF}{{\bf F}}
\newcommand{\bfS}{{\bf S}}
\newcommand{\bfC}{{\bf C}}
\newcommand{\bfI}{{\bf I}}
\newcommand{\bfO}{{\bf O}}
\newcommand{\bfU}{{\bf U}}
\newcommand{\bfD}{{\bf D}}
\newcommand{\bfY}{{\bf Y}}
\newcommand{\bSig}{{\bf \Sigma}}
\newcommand{\test}{\stackrel{<}{>}}
\newcommand{\zmk}{{\bf Z}_{m,k}}
\newcommand{\zlk}{{\bf Z}_{l,k}}
\newcommand{\zm}{{\bf Z}_{m}}
\newcommand{\ssq}{\sigma^{2}}
\newcommand{\dint}{{\displaystyle \int}}
\newcommand{\ds}{\displaystyle }
\newtheorem{theorem}{Theorem}
\newcommand{\postscript}[2]{ \begin{center}
    \includegraphics*[width=3.5in,height=#1]{#2.eps}
    \end{center} }

\newtheorem{identity}{Identity}
\newtheorem{hypothesis}{Hypothesis}
\newcommand{\mathtiny}[1]{\mbox{\tiny$#1$}}

\maketitle

\begin{abstract}
In this paper, it is shown that an auto-encoder
using optimal reconstruction
significantly outperforms a conventional auto-encoder. 
Optimal reconstruction uses the conditional
mean of the input given the features, under a maximum entropy
prior distribution. The optimal reconstruction network,
which is called  deterministic projected belied network (D-PBN),
resembles a standard reconstruction network, but with special non-linearities
that mist be iteratively solved.
The method, which can be seen as a generalization of maximum entropy image reconstruction, extends to multiple layers.
In experiments, 
mean square reconstruction error reduced by up to a factor of
two.  The performance improvement diminishes for deeper networks,
or for input data with unconstrained values (Gaussian assumption).
\end{abstract}

\section{Introduction}

\subsection{Problem Definition and Background}
Despite the many approches to training,
virtually all neural network layers  begin with linear feature extraction.
Although there are some exceptions, it is the
dimension-reducing linear transformation that is key
to extracting the useful information in a useable form.
Therefore, it is important to study 
linear dimension-reducing transformations
with optimal reconstruction in mind.
Without knowing the ultimate application of feature extraction,
it is reasonable to define optimality in terms of the ability
to reconstruct the input data of a layer from
the output of that layer.  This criterion is
especially relevant for an auto-encoder. 
Let $\bfx\in\mathbb{R}^N$ be the input data vector
and let $\bfz={\bf W}^\prime \bfx \in\mathbb{R}^M$ 
be the extracted feature where $M<N$,
where $'$ is the transpose operator. Matrix ${\bf W}$ is the
$N\times M$ weight matrix.  
The bias and activation function are ignored for the time being.

Feature inversion 
is a well studied topic \cite{Boucheron08,milner2002speech,Chazan}.
Since the problem is underdetermined, regularization
is required to limit the norm of the estimated $\bfx$ \cite{Boucheron08}.
An example is image reconstruction in which regularization is achieved by seeking
the input data with highest entropy \cite{Wernecke77,Wei87}.

Taking a probabilistic point of view, 
we seek an estimate of $\bfx$ conditioned on knowing $\bfz$.
The conditional mean estimate, written $\mathbb{E}(\bfx|\bfz)$
is the minimum mean-square estimator (MMSE) \cite{KayEst},
however a prior distribution is required.  The
principle of maximum entropy proposes that with no other prior knowledge,
the prior distribution should be the one with highest
entropy among all distributions meeting the given
constraints \cite{Jaynes82}.  Typical constraints for 
$\bfx$ and its distribution come from the given data range,
(if the elements of $\bfx$ are positive, constrained to
$[0, \; 1]$, or unconstrained), and any assumptions of mean or variance.
Recently, it was shown how to solve for the conditional mean 
for data ranges and maximum entropy priors suitable for machine learning applications.
The result is an auto-encoder called
 deterministic projected belief network (D-PBN)
\cite{BagEusipcoPBN,BagIcasspPBN,BagPBNEUSIPCO2019,BagPBNEUSIPCO2020}.

\subsection{Paper Summary and Contributions}
In this paper, we review the mathematical
background for the D-PBN, then conduct
experiments comparing D-PBN with auto-encoders based on conventional
reconstruction networks including the variational auto-encoder (VAE).
The main contributions of this work are the experimental results
which show that the D-PBN significantly improves over
auto-encoders with conventional reconstruction networks.
This improvement depends on the depth of the network and
the type of data range. In particular, when the Gaussian assumption
can be made, the D-PBN is equivalent to conventional
auto-encoders.  However, for input data constrained to be positive, or to the
range $[0,\;1]$, the D-PBN proves to be superior, although this improvement
diminishes for deeper networks.



\section{Mathematical Results}

\subsection{Data Ranges and MaxEnt Priors}
In order to solve for the  conditional mean, we need
a prior distribution, and this depends on the assumed
range of the input data, denoted by $\mathbb{X}$.
As before, let $\bfx\in\mathbb{R}^N$ be the input data vector
and let $\bfz={\bf W}^\prime \bfx \in\mathbb{R}^M$ 
be the extracted feature where $M<N$.  
The task at hand is to arrive at a posterior distribution
$p(\bfx|\bfz)$, then determine $\mathbb{E}(\bfx|\bfz)$.  
For most purposes, it is sufficient to consider three
data ranges: the full range $\mathbb{R}^N$, the positive quadrant denoted
$\mathbb{P}^N$ for which $0<x_i$, and the unit hypercube
denoted $\mathbb{U}^N$ for which $0<x_i<1$. 

Maximum entropy (MaxEnt) distributions are
generally of the exponential class \cite{Kapur}.
For these data ranges, the prior distributions 
are in the class of distributions given by
\vspace{-.1in}
\beq
p_e(\bfx; {\bf a},b) = \prod_{i=1}^N p_e(x_i; a_i,b),
\label{pedef}
\vspace{-.1in}
\eeq
where
\vspace{-.1in}
\beq
p_e(x;a,b) =\frac{1}{K(a,b)} e^{a x + b x^2}.
\label{exppr}
\vspace{-.1in}
\eeq
Since the entropy of a distribution increases with variance,
entropy can grow without bounds in the ranges $\mathbb{R}^N$ and $\mathbb{P}^N$
unless the variance is constrained\footnote{Technically, 
only the mean needs to be constrained for $\mathbb{P}^N$,
but the resulting models are not as useful for neural networks.},
therefore we set $b=\frac{1}{2}$ for these data ranges
to arrive at well-known canonical prior distributions.
For $\mathbb{U}^N$, we can set $b=0$ and $a=0$, resulting
in the uniform distribution.
For a given quadratic parameter $b$, the univariate distribution (\ref{exppr})
has a mean that is a function of $a$, denoted by
\vspace{-.1in}
\beq
\lambda(a)=\int x \; p_e(x;a,b) {\rm d} x.
\label{actdef}
\vspace{-.1in}
\eeq
This function, which we call the MaxEnt activation function,
 is central to our approach.  

The MaxEnt prior distributions, which are derived from
(\ref{pedef}) by setting $a$ and $b$ are listed in Table \ref{tab1v} 
for the three data ranges $\mathbb{R}^N$, $\mathbb{P}^N$, and $\mathbb{U}^N$.
These are Gaussian, Truncated Gaussian (TG) and uniform, respectively.
The $\mathbb{P}^N$ data range also admits the exponential 
prior, which is added for completeness, but is not as useful
for machine learning.
The MaxEnt activation functions are also shown.
\begin{table}
\begin{center}
 \begin{tabular}{|l|l|l|l|}
\hline
         $\mathbb{X}$ & $a,b$ &  MaxEnt Prior $p_e(x;a,b)$  & $\lambda(a)$ \\
 \hline
         $\mathbb{R}^N$  & $ 0, -\frac{1}{2}$  & $\phi(x,0,\sigma^2)$  (Gauss) & $\sigma^2 a$  (Linear)\\
 \hline
         $\mathbb{P}^N$  & $ 0, -\frac{1}{2}$  & $2 \phi(x,0,\sigma^2)$ (TG) & $\sigma^2 a + \sigma \frac{\phi(\sigma a;0,1)}{ \Phi(\sigma a)}$  (TG)\\
 \hline
         $\mathbb{P}^N$  & $ -1, 0$  & $e^{-x}$ (exponential) & $-\frac{1}{ a}$ \\
 \hline
         $\mathbb{U}^N$ & 0,0&  $1$  (Uniform) & $\frac{e^{ a}}{e^{ a} - 1}-\frac{1}{ a}$ (TED)  \\
 \hline
\end{tabular}
\end{center}
\caption{MaxEnt priors and activation functions as a function of input data range.
        TG=``Trunc. Gauss.". TED=``Trunc. Expon. Distr". Notation:
$\phi(x;a,\sigma^2)= (2\pi\sigma^2)^{-1/2} \; e^{-(x-a)^2/(2\sigma^2)},$
and $\Phi(\;)$ is the cumulative distribution of the standard normal distribution
$\Phi\left( x\right)  \defined \int_{-\infty}^x \phi\left(x,0,1\right).$
}
\label{tab1v}
\vspace{-.2in}
\end{table}

\subsection{Posterior Distribution}
Now that the MaxEnt prior has been defined, we can proceed to find $p(\bfx|\bfz)$.
Clearly, given that $\bfz$ has been observed, $\bfx$ must exist on the manifold
${\cal M}_z=\{\bfx : {\bf W}^\prime \bfx=\bfz\}.$
The posterior $p(\bfx|\bfz)$ is not a proper distribution on $\mathbb{X}$ because
it has support only on the set ${\cal M}_z$ which has zero volume. As a result,
the conditional mean that we seek, denoted by $\mathbb{E}(\bfx|\bfz)$ 
cannot be written in closed form.
The method of {\it surrogate density} \cite{BagUMS}
proposes that a distribution of the form (\ref{pedef}),
which is a proper distribution on $\mathbb{X}$,
approaches the true posterior, so has the same mean,
and $\mathbb{E}(\bfx|\bfz)$ can be derived from it.
The conditional mean 
asymptotically (for large $N$) approaches
$$\bar{\bfx}_z \defined \mathbb{E}(\bfx|\bfz)=\lambda\left(
{\bf W} \bfh \right),$$
where $\lambda(\;)$ is from (\ref{actdef})
and operates element-wise on a vector,
and $\bfh$ is the solution to:
\beq
{\bf W}^\prime \lambda\left( {\bf W} \bfh \right) = \bfz.
\label{uuu}
\eeq
The solution $\bfh$ always exists when
$\bfz={\bf W}^\prime \bfx$ for some $\bfx\in\mathbb{X}$ 
and is also known as the saddle point \cite{BagIcasspPBN}.
The MaxEnt activation, defined in (\ref{actdef}), depends on $\mathbb{X}$ and the corresponding MaxEnt reference distribution, which takes the form (\ref{pedef})
for a specific choice of $a,b$.
%
%
The MaxEnt reference hypotheses and the corresponding activation
functions $\lambda(\;)$ are listed in Table \ref{tab1v}. 
Note that the TED and TG activation functions are similar in behavior
to the {\it sigmoid} and {\it softplus} activations in common use,
respectively (Figure 2 in \cite{BagIcasspPBN}).

\subsection{Special Non-linearity}
\begin{figure}[h!]
  \begin{center}
    \includegraphics[width=3.5in]{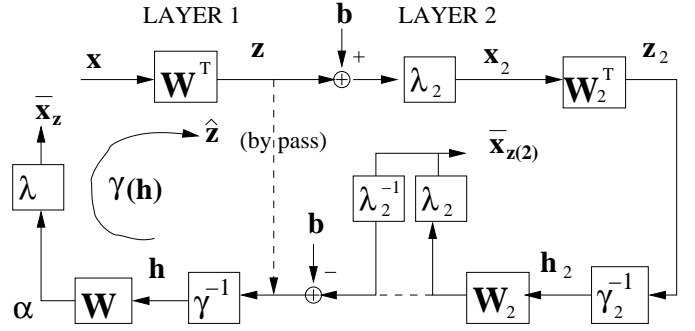}
  \caption{Block diagram of 1 or 2-layer D-PBN (adapted from \cite{BagIcasspPBN}).}
  \label{asy}
  \end{center}
\end{figure}

Figure \ref{asy} illustrates a 1 or 2-layer D-PBN.
A 1-layer network is created by following the
shortcut denoted by ``by pass".
Although the MaxEnt activations $\lambda(\;)$ 
used in the forward path (see Fig. \ref{asy}, top)
are similar to widely used activations and are applied element-wise,
a special nonlinearity is used in the reconstruction path
and does not operate element-wise.  For simplicity, define the function
$\hat{\bfz}=\gamma(\bfh) =  {\bf W}^\prime \lambda\left( {\bf W} \bfh \right),$
which reconstructs $\bfz$ from an intermediate variable $\bfh$.
This function is illustrated by the circular path in Figure \ref{asy}.
Solving (\ref{uuu}) for $\bfh$ can be written $\bfh=\gamma^{-1}(\bfz)$
and requires Newton's method \cite{BagUMS}.

\subsection{Multiple Layers}
\label{multilay}
A D-PBN can be extended to any number of layers.  Figure \ref{asy} shows a 2-layer D-PBN.
In the forward path (top), a bias is added to the output of the 
linear transformation of the first layer $\bfz$,
and an activation function is applied $\lambda_2(\;)$.
This activation function must be selected from Table \ref{tab1v},
which will define the data range for the second layer.
Note that these activation functions have similar behavior to
activation functions in common use, {\it linear}, {\it sigmoid}, {\it softplus}.
In fact, the TG activation can be made to work like {\it ReLu}
by adjusting the variance parameter $\sigma^2$.

Let $\bfz_2$ be the output of second layer linear transformation
and  $\bar{\bfx}_{z(2)}$ be the reconstructed input to the second layer.
In order to proceed, $\bar{\bfx}_{z(2)}$ must be converted so that it is
compatible with $\bfz$, the output of the linear transformation of the first
layer. To do this, it is necessary to invert the activation function
at the output of the first layer, then subtract the
bias vector (See Fig.  \ref{asy}).  Although all MaxEnt activation
functions are theoretically invertible, inversion
can be avoided if the activation function at the output of the previous layer
is the same as the MaxEnt activation function
of the current layer \cite{BagPBNEUSIPCO2019}.
To see this, note that $\bar{\bfx}_z =\lambda\left( {\bf W} \bfh \right).$
It is then only necessary to take the quantity ${\bf W} \bfh$,
then subtract the bias of the prevous layer
(see dotted line on bottom of Fig.  \ref{asy}). The resulting quantity
is used in place of $\bfz$ for the previous layer (See Fig. \ref{asy}).
This process repeats until the visible data has been generated.

Note that in multi-layer PBNs, there is a chance that
no solution $\bfh$ to equation (\ref{uuu}) can be found and sampling fails,
but the problem can be dealt with as explained in Section \ref{sampfail}.

\subsection{D-PBN as an auto-encoder}
\begin{figure}[h!]
  \begin{center}
    \includegraphics[width=3.5in]{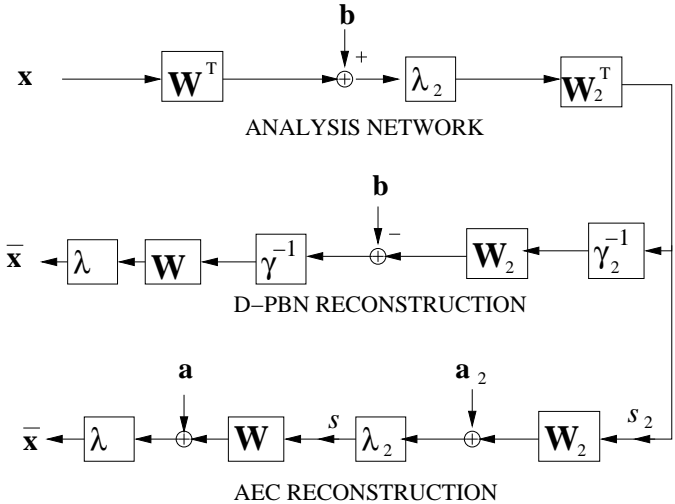}
  \caption{Block diagram of the 2-layer networks showing analysis network (top), 
        D-PBN reconstruction network (center), and conventional
   auto-encoder reconstruction network (bottom). }
  \label{asy3}
  \end{center}
\end{figure}

The deterministic PBN (D-PBN) is trained like an auto-encoder
to minimize visible data reconstruction error.
Figure \ref{asy3} is a simplification of Figure \ref{asy} and includes
a standard reconstruction network for comparison.
As illustrated in Figure \ref{asy3}, the forward path
is a standard multi-layer perceptron with conventional activation functions
(conventional in the sense that they are monotonic increasing and operate element-wise).
Whereas the standard auto-encoder (AEC) uses a perceptron
network for reconstruction, the D-PBN reconstruction network uses a special
activation function written $\bfh=\gamma^{-1}(\bfz)$.
In effect, the D-PBN is a special type of auto-encoder
with tied analysis and reconstruction weights
(i.e. the decoder network is the transpose of the encoder network).
It has been shown that the D-PBN reconstruction network is the dual of the encoder network \cite{BagPBN}.
This fact has the potential to improve generalization
due to reduction of parameter count and because
the decoder is in a sense ``perfectly matched" to the encoder.

\subsection{Relationship to MaxEnt Image Reconstruction}
\begin{figure}[h!]
  \begin{center}
  \includegraphics[width=1.1in,height=1.1in]{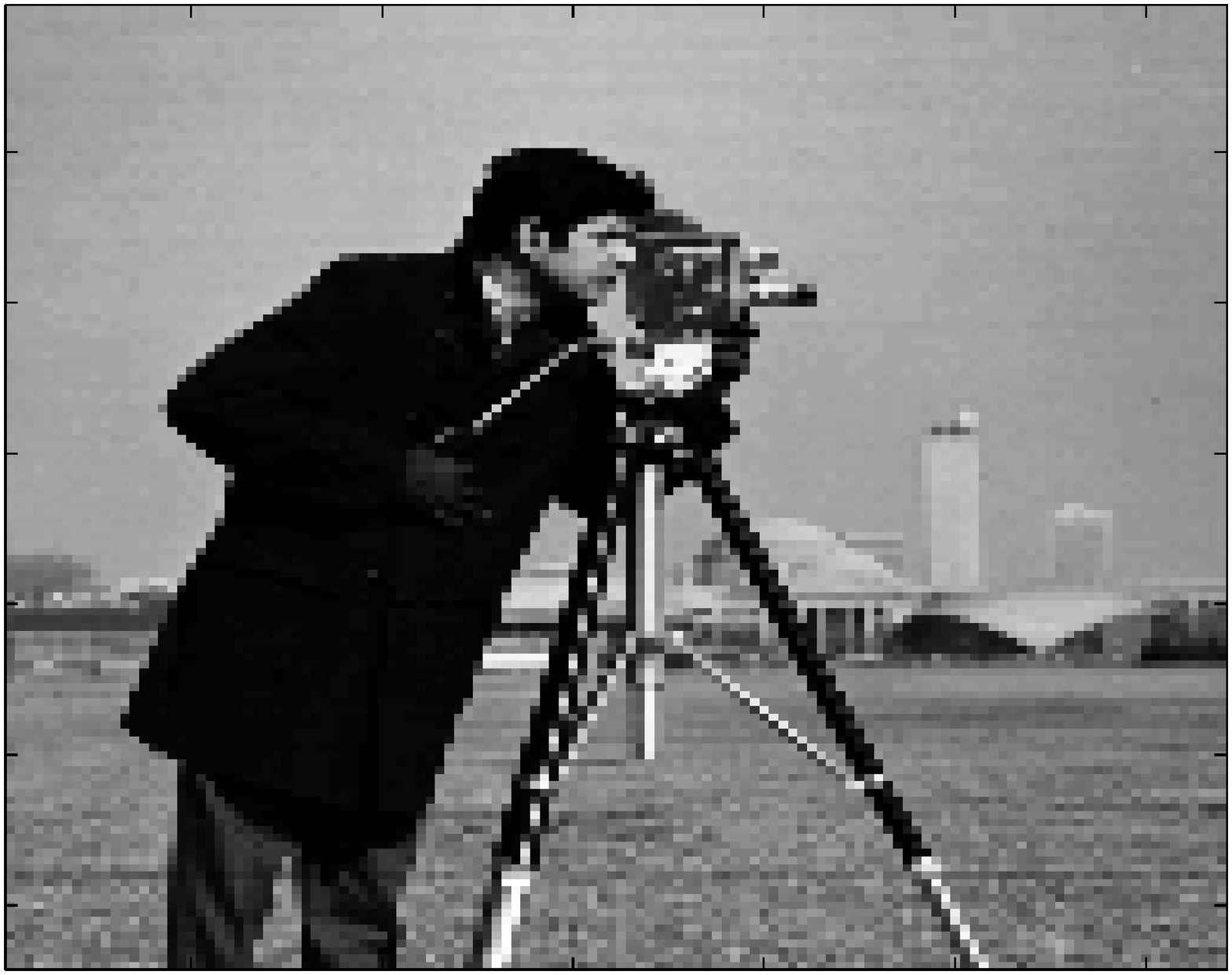}
  \includegraphics[width=1.1in,height=1.1in]{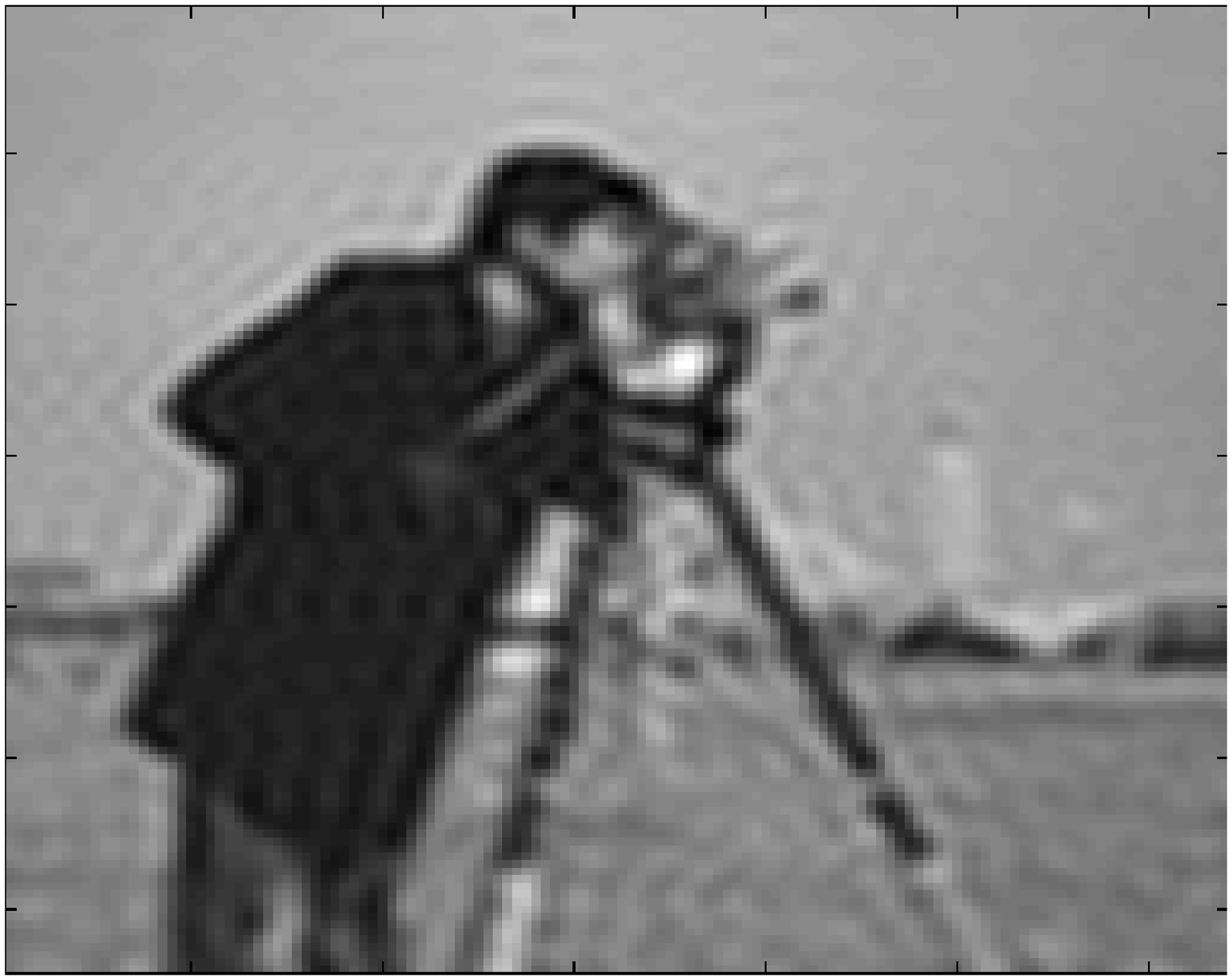}
  \includegraphics[width=1.1in,height=1.1in]{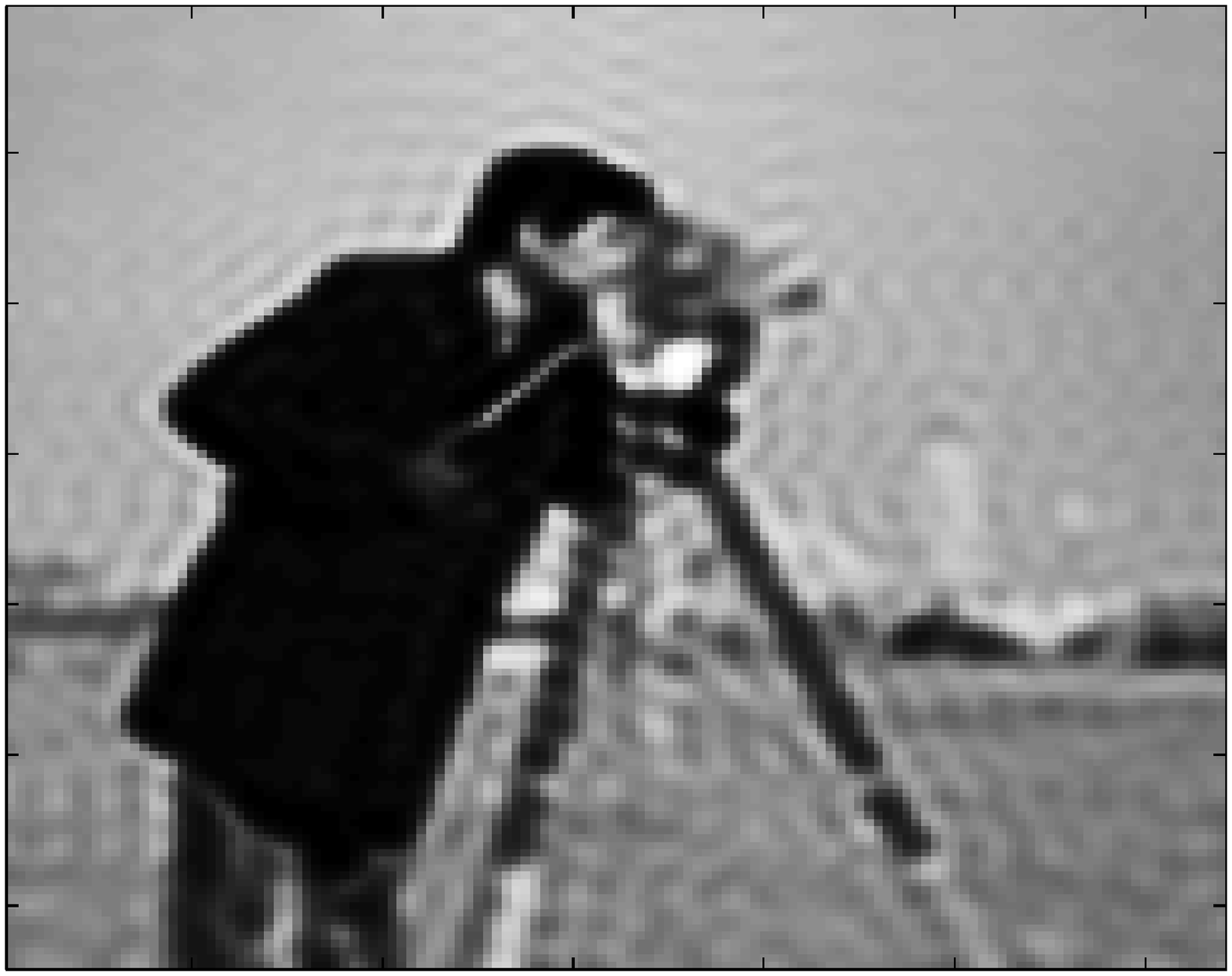}
  \caption{Example of MaxEnt Image Reconstruction (reprinted from \cite{BagUMS}).
Left: original $128\times128$ image. Center: reconstructed from $48\times48$ DCT coefficients.
Right: MaxEnt reconstruction on $[0, 1]$ with uniform prior.}
  \label{imgrc}
  \end{center}
  \vspace{-.15in}
\end{figure}

Our approach, which we call alternatively MaxEnt auto-encoding or
deterministic projected belief network (D-PBN), is a generalization
of MaxEnt image reconstruction.  The approach generalizes the
traditional approach to different data ranges and MaxEnt prior
distributions.  A table of data ranges and
prior distributions is given in Table \ref{tab1v}.
It was shown that traditional MaxEnt image reconstruction \cite{Wernecke77,Wei87},
is equivalent to our approach for the data range $[0, \infty]$ using the exponential
prior \cite{BagUMS}.  An example of image reconstruction
with the uniform prior on the range $[0, 1]$ is shown in Figure \ref{imgrc}.  
A sharper image results for the same feature dimension.
In this paper, we generalize further
to the truncated Gaussian assumption (See Table \ref{tab1v}) as well as to the
multiple layers.

\section{Experiments}


\subsection{Training and Implementation}
\label{sampfail}
The D-PBN was trained by minimizing the average reconstruction
error.  The network was trained with the PBN Toolkit \cite{PBNTk},
using Python-based Theano \cite{Theano2010} as a 
framework and GPU hardware.  Training a D-PBN is typically one order
of magnitude slower than a standard AEC. Derivatives needed for stochastic gradient
training were generated by the symbolic
processing of Theano. Note that solving (\ref{uuu})
is iterative, so derivatives are based on analysis of 
the last iterations.

As explained in Section \ref{multilay}, there is a chance
that no visible data sample can be reconstructed
from a given network output feature. The {\it sampling efficiency}
is the probability of successful reconstruction.
To increase sampling efficiency, the network can be initialized as a 
stacked restricted Boltzmann machine \cite{WellingHinton04,HintonDeep06}, 
then trained as a standard stochastic PBN \cite{BagPBN}.
Once sampling efficiency is larger then about 0.5, training
of the D-PBN can proceed as long as contributions of the
failed samples are removed from the derivatives.
A failed sample is detected when the error between 
$\hat{\bfz}=\gamma(\bfh)$ and $\bfz$ (See Fig. \ref{asy}) 
is much larger then machine precision.
During training, the sampling efficiency then rapidly approaches 1.
In all the experiments we conducted, not a single failed
sample was detected either for training or testing data
for fully-trained D-PBN networks.

\subsection{Data Set}
\label{datsec}
In the experiments, we used the MNIST
handwritten character data set.
The data consists of $28\times 28$ sample images, with dimension $N=784$.
There are 6000 training samples and 1000 testing samples
of each character.
In order to adapt the data to any of the data ranges $\mathbb{R}^N$,
$\mathbb{P}^N$, or $\mathbb{U}^N$, 
we first ``gaussianified" the data by adding
dither to the quantized data\footnote{For pixel values
above 0.5, a small exponential-distributed random value was subtracted,
but for pixel values below 0.5, a similar  random value was
added.}, then applied the inverse sigmoid function, 
resulting in a smooth Gaussian-like distribution  in the range -10 and 10.
This ``gaussianified" data was then passed through a MaxEnt activation function
to create the desired data range.

For multi-layer experiments, just three characters ``3", ``8", and ``9" were chosen
and only 500 samples of each character were used to train.
This reduced computational complexity and also highlighted
the generalization capability of D-PBN by reducing the training set.


\subsection{Single-Layer Experiments}
In the 1-layer experiment, the task was to encode samples of the three MNIST characters
in each of the data ranges ($\mathbb{R}$, $\mathbb{P}$, and $\mathbb{U}$)
to a dimension of $M=24$ using a single linear transformation, 
then reconstruct the visible data.  We compare the D-PBN
with a restricted Boltzmann machine (RBM) \cite{WellingHinton04}
 and a conventional auto-encoder (AEC) using the same 1-layer network.
For RBM, and AEC, the same weight matrix was used for analysis and reconstruction
(tied weights), although a separate
scale factor (see $s$ in Figure \ref{asy3}) was allowed for reconstruction
to account for different scaling, and separate
reconstruction bias was used.
For all methods, a linear output activation 
was assumed (i.e. no activation function). The reconstruction activation
function was matched to the data range:
linear for $\mathbb{R}$, TG for $\mathbb{P}$, 
and TED for $\mathbb{U}$ (See Table \ref{tab1v}).
The auto-encoders were evaluated by subjective comparison
of the reconstructed characters and
quantitative comparison using reconstruction error.

\begin{figure*}[h!]
  \begin{center}
  \includegraphics[width=6.5in,height=2.2in]{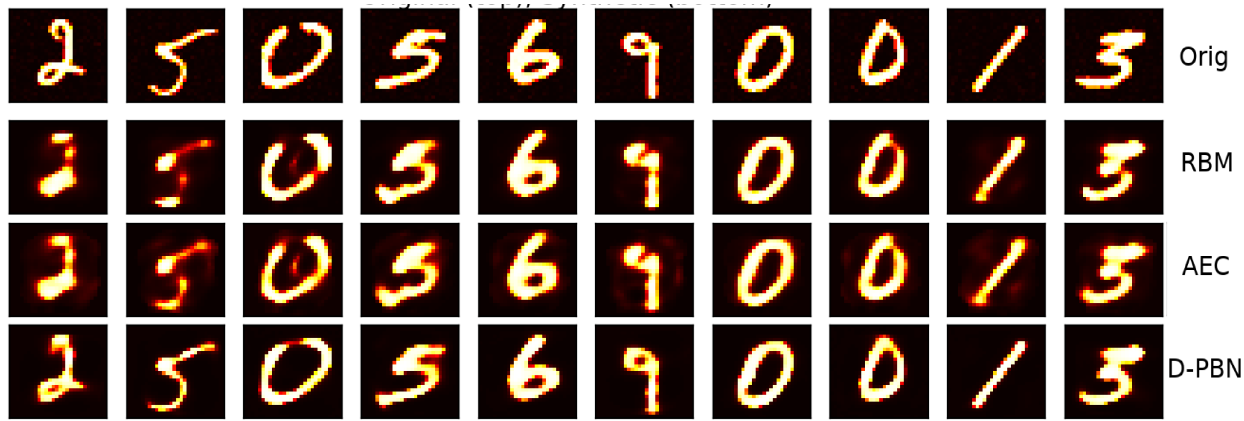}
  \caption{Reconstructed MNIST characters for 1-layer
network on data range $\mathbb{U}^N$. All models reconstructed
data from a 24-dimensional feature.}
  \label{reconU}
  \end{center}
\vspace{-.15in}
\end{figure*}

Table \ref{reconU} lists the mean-square reconstruction error
for three network types and three data ranges.
For $\mathbb{R}^N$, there is almost no difference between
the methods, which can be explained by the fact that the
conditional mean estimator is just least-squares which
can be implemented by a linear transformation. Therefore,
all methods can approximate the true conditional mean estimator.
There is also very little difference in MSE for training
and test data, a result of low network complexity and
sufficient training data size.

For $\mathbb{P}^N$, the performance of the three methods varies significantly.
Note that the input data is differently scaled for the three
data ranges, which explains the large differences in
MSE, but for a given data range, all three auto-encoders
operated on the same data.
D-PBN shows a  35\% reduction in MSE compared to AEC. 
For $\mathbb{U}^N$, the performance of the three methods becomes
even more varied, with D-PBN showing a factor of 2 reduction in MSE compared to AEC. 
One can conclude from this that the benefit of D-PBN is related to the
``non-Gaussianity" of the data.  In fact, for the Gaussian assumption, there is no benefit
at all, and for data constrained to $[0, \; 1]$, the benefit
is very significant.  Figure \ref{reconU} shows the reconstructions
for a random-sampling of ten characters.
The quality of the D-PBN reconstruction is clear to see, especially for
thinly drawn characters.
\begin{table}[htb]
\begin{center}
 \begin{tabular}{|l|l|l|l|l|l|l|}
\hline
 &  \multicolumn{2}{|c|}{RBM} &  
      \multicolumn{2}{|c|}{AEC}  
      & \multicolumn{2}{|c|}{D-PBN} \\
\hline
Data Range $\mathbb{X}$ & Train & Test & Train & Test & Train & Test \\
\hline
$\mathbb{R}^N$ & 3.09 & 3.10 & 3.09 & 3.11 & 3.09 & 3.10 \\
 \hline
$\mathbb{P}^N$ & .896 & .911 & .688 & .696 & .5087 & .5145\\
 \hline
$\mathbb{U}^N$ & .0172 & .0173 & .0152 & .0154 & .00725 & .00734 \\
 \hline
\end{tabular}
\end{center}
	\caption{Mean-square pixel reconstruction error 
for 1-layer networks as a function of reconstruction model and data range.}
	\label{tab2}
\vspace{-.15in}
\end{table}

\subsection{Multi-Layer Experiments}
Given the clear advantage of D-PBN data reconstruction in the single-layer
experiment, it is relevant to ask if this holds up in a milti-layer network. 
After all, auto-encoders typically use non-linear activation functions
in intermediate layers, with data supports of
$\mathbb{U}^N$ for {\it sigmoid} and $\mathbb{P}^N$ for {\it softplus}
or {\it ReLu}, and it is these data ranges where the D-PBN shows advantage.

Experiments were carried out with two and three layers.
Node counts for the 2-layer network were $48,24$, 
and $48,24,12$ for the 3-layer networks.  The layer output activation functions
were chosen to correspond to the
input data range, so that all layers had the same MaxEnt activation.
A linear activation was used for the output of the last layer.
For the multi-layer experiments, we used just a subset of the
MNIST data set, as explained in Section \ref{datsec}.

Instead of using RBM as a third model
for comparison, we used a variational auto-encoder (VAE)
which is based on a probabilistic model of the data, and
has gained popularity for unsupervised learning
tasks \cite{doersch2016tutorial}.
Although the VAE is technically not an auto-encoder (it is a probabilistic
data model), an auto-encoder is integral to its operation.
The VAE cost function employs a Kullback-Leibler divergence (KLD)
term which provides good regularization effect
\cite{odaibo2019tutorial,pmlr-v32-rezende14}.

Table \ref{tab3} lists the MSE for the three 
auto-encoder networks VAE, AEC, and D-PBN.
Results for $\mathbb{R}^N$ are not included because it is a trivial case,
as seen in the 1-layer results.
On $\mathbb{P}^N$, the VAE has higher MSE than the AEC
on training data, but lower on testing data, a result of
regularization effect of the KLD cost function.
The D-PBN is superior, for both training and testing data.

On data range $\mathbb{U}^N$, the VAE has higher MSE than the AEC
on both training and testing data, an indication of the
very high regularization effect. Recall that
the VAE is primarily a statistical model.
It is interesting to note that even though the
D-PBN had lowest MSE on the training data, it also has better
generalization performance compared to the other models.
To see if additional regularization could
improve the AEC, the experiment was re-run with L2 regularization,
which was optimized for generalization performance.
Even with optimal L2 regularization, the D-PBN performance
was best.
This indicates that D-PBN has a built-in regularization effect,
probably due to the probabilistic basis for the model.

\begin{table}[htb]
\begin{center}
 \begin{tabular}{|l|l|l|l|l|l|l|l|}
\hline
&
 &  \multicolumn{2}{|c|}{VAE}   
 &     \multicolumn{2}{|c|}{AEC}  
      & \multicolumn{2}{|c|}{D-PBN} \\
\hline
Data Range & L2 & Train & Test & Train & Test & Train & Test \\
\hline
$\mathbb{P}^N$ & 0 & .580 & .780 & .500 & .817 & .441 & .517 \\
\hline
$\mathbb{P}^N$ &  5e-3 &  &  & .552 & .637  &  &  \\
 \hline
$\mathbb{U}^N$ & 0 & .0236 & .0254 & .0104 & .0112 & .00773 & .0087\\
 \hline
\end{tabular}
\end{center}
	\caption{Mean-square pixel reconstruction error 
for 2-layer networks as a function of reconstruction model and data range.}
	\label{tab3}
\vspace{-.15in}
\end{table}

Table \ref{tab4} shows  the results for the 3-layer experiments.
The D-PBN had significantly lower reconstruction error
only for $\mathbb{U}^N$. For $\mathbb{P}^N$,
it was the same as the AEC.

\begin{table}[htb]
\begin{center}
 \begin{tabular}{|l|l|l|l|l|l|l|l|}
\hline
&
 &  \multicolumn{2}{|c|}{VAE}   
 &     \multicolumn{2}{|c|}{AEC}  
      & \multicolumn{2}{|c|}{D-PBN} \\
\hline
Data Range & L2 & Train & Test & Train & Test & Train & Test \\
\hline
$\mathbb{P}^N$ & 0 & .7431 & 1.51 & .651 & .867 & .7153 & .8665 \\
\hline
$\mathbb{U}^N$ & 0 & .216 & .224 & .0255 & .0233 & .0143 & .0152\\
 \hline
\end{tabular}
\end{center}
	\caption{Mean-square pixel reconstruction error 
for 3-layer networks as a function of reconstruction model and data range.}
	\label{tab4}
\vspace{-.2in}
\end{table}

\section{Discussion, Future Work,  and Conclusions}
\begin{figure}[h!]
  \begin{center}
  \includegraphics[width=3.5in, height=1.3in]{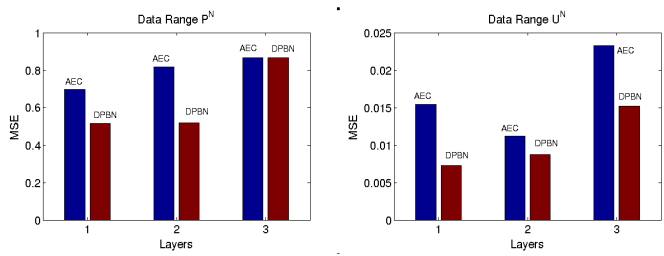}
  \caption{Summary of results by layer.}
  \label{sumry}
  \end{center}
  \vspace{-.15in}
\end{figure}
Results are summarized in Figure \ref{sumry},
which shows MSE as a function of layer for AEC and D-PBN.
For data range $\mathbb{P}^N$, with data in $[0, \; \infty]$,
the D-PBN results in lower MSE, but the improvement
vanishes for more than 2 layers.
For data range $\mathbb{U}^N$, with data in $[0, \; 1]$,
the the improvement of D-PBN is high as a factor of 2,
and continues to be significant at 3 layers.
This indicates that the D-PBN is preferable
for auto-encoding in shallow networks and/or
when data is constrained to $[0, \; 1]$.
The sigmoid activation, which produces data in the range
$[0, \; 1]$, is no longer widely-used and has been
replaced by ReLU and softplus, which carry more amplitude information
and are not affected by the problem of vanishing gradients.
However, in light of the above results, it may be worth another look.
As future work, this must be tested on other data sets,
and on deeper networks.

\bibliographystyle{ieeetr}
\bibliography{ppt}
\end{document}